%% file: tacl.tex
\documentclass[11pt,letterpaper]{article}
\usepackage[utf8]{inputenc} % allow utf-8 input
\usepackage[T1]{fontenc} % use 8-bit T1 fonts
\usepackage[letterpaper]{geometry}
\usepackage{acl2012}
\usepackage{times}
\usepackage{latexsym}
\usepackage{multirow}
\usepackage{url}
\makeatletter
\newcommand{\@BIBLABEL}{\@emptybiblabel}
\newcommand{\@emptybiblabel}[1]{}
\makeatother
\usepackage[hidelinks]{hyperref}
\usepackage{booktabs}
\usepackage{graphicx}
\usepackage{caption}
\usepackage{subcaption}
\usepackage{amssymb}
\usepackage{amsmath}
\usepackage{soul}
\usepackage{color}
\usepackage{relsize}

\title{Enriching Word Vectors with Subword Information}

\author{Piotr Bojanowski\thanks{\ \ The two first authors contributed equally.} \and Edouard Grave\footnotemark[1] \and Armand Joulin \and Tomas Mikolov \\
  Facebook AI Research \\
  \texttt{\{bojanowski,egrave,ajoulin,tmikolov\}@fb.com}
}

\date{}

\begin{document}
\maketitle

\begin{abstract}
Continuous word representations, trained on large unlabeled corpora are useful for many natural language processing tasks.
Popular models that learn such representations  ignore the morphology of words, by assigning a distinct vector to each word.
This is a limitation, especially for languages with large vocabularies and many rare words.
In this paper, we propose a new approach based on the skipgram model, where each word is represented as a bag of character $n$-grams.
A vector representation is associated to each character $n$-gram; words being represented as the sum of these representations.
Our method is fast, allowing to train models on large corpora quickly and allows us to compute word representations for words that did not appear in the training data.
We evaluate our word representations on nine different languages, both on word similarity and analogy tasks.
By comparing to recently proposed morphological word representations, we show that our vectors achieve state-of-the-art performance on these tasks.
\end{abstract}

\section{Introduction}

Learning continuous representations of words has a long history in natural language processing~\cite{rumelhart1988learning}.
These representations are typically derived from large unlabeled corpora using co-occurrence statistics~\cite{deerwester1990indexing,schutze1992dimensions,lund1996producing}.
A large body of work, known as distributional semantics, has studied the properties of these methods~\cite{turney2010frequency,baroni2010distributional}.
In the neural network community, \newcite{collobert2008unified} proposed to learn word embeddings using a feedforward neural network, by predicting a word based on the two words on the left and two words on the right.
More recently, \newcite{mikolov2013distributed} proposed simple log-bilinear models to learn continuous representations of words on very large corpora efficiently. 

%\begin{table}[t]
%  \centering
%  \begin{tabular}{lccc}
%    \toprule
%    & singular & plural \\
%    \midrule
%    first  & écrirai & écrirons \\
%    second & écriras & écrirez  \\
%    third  & écrira  & écriront \\
%    \bottomrule
%  \end{tabular}
%  \caption{Example of conjugation of the French verb \emph{écrire} (to write, future tense).}
%\end{table}

Most of these techniques represent each word of the vocabulary by a distinct vector, without parameter sharing.
In particular, they ignore the internal structure of words, which is an important limitation for morphologically rich languages, such as Turkish or Finnish.
For example, in French or Spanish, most verbs have more than forty different inflected forms, while the Finnish language has fifteen cases for nouns.
These languages contain many word forms that occur rarely (or not at all) in the training corpus, making it difficult to learn good word representations.
Because many word formations follow rules, it is possible to improve vector representations for morphologically rich languages by using character level information.

In this paper, we propose to learn representations for character $n$-grams, and to represent words as the sum of the $n$-gram vectors.
Our main contribution is to introduce an extension of the continuous skipgram model~\cite{mikolov2013distributed}, which takes into account subword information.
We evaluate this model on nine languages exhibiting different morphologies, showing the benefit of our approach.

\input{relatedwork}

\input{model}

\input{experiments}

\section{Conclusion}

In this paper, we investigate a simple method to learn word representations by taking into account subword information.
Our approach, which incorporates character $n$-grams into the skipgram model, is related to an idea that was introduced by \newcite{schutze1993word}.
Because of its simplicity, our model trains fast and does not require any preprocessing or supervision.
We show that our model outperforms baselines that do not take into account subword information, as well as methods relying on morphological analysis.
We will open source the implementation of our model, in order to facilitate comparison of future work on learning subword representations.

% \clearpage

\subsection*{Acknowledgements}
We thank Marco Baroni, Hinrich Schütze and the anonymous reviewers for their insightful comments.

\bibliography{emnlp2016}
\bibliographystyle{acl2012}

\end{document}

%% file: relatedwork.tex
\section{Related work}

\paragraph{Morphological word representations.}
In recent years, many methods have been proposed to incorporate morphological information into word representations.
To model rare words better, \newcite{alexandrescu2006factored} introduced factored neural language models, where words are represented as sets of features.
These features might include morphological information, and this technique was succesfully applied to morphologically rich languages, such as Turkish~\cite{sak2010morphology}.
Recently, several works have proposed different composition functions to derive representations of words from morphemes~\cite{lazaridou2013compositional,luong2013better,botha2014compositional,qiu2014colearning}.
These different approaches rely on a morphological decomposition of words, while ours does not.
Similarly, \newcite{chen2015joint} introduced a method to jointly learn embeddings for Chinese words and characters.
\newcite{cui2015knet} proposed to constrain morphologically similar words to have similar representations.
\newcite{soricut2015unsupervised} described a method to learn vector representations of morphological transformations, allowing to obtain representations for unseen words by applying these rules.
Word representations trained on morphologically annotated data were introduced by \newcite{cotterell2015morphological}.
Closest to our approach, \newcite{schutze1993word} learned representations of character four-grams through singular value decomposition, and derived representations for words by summing the four-grams representations.
Very recently, \newcite{wieting2016charagram} also proposed to represent words using character $n$-gram count vectors.
However, the objective function used to learn these representations is based on paraphrase pairs, while our model can be trained on any text corpus.

\paragraph{Character level features for NLP.}
Another area of research closely related to our work are character-level models for natural language processing.
These models discard the segmentation into words and aim at learning language representations directly from characters.
A first class of such models are recurrent neural networks, applied to language modeling~\cite{mikolov2012subword,sutskever2011generating,graves2013generating,bojanowski2015alternative}, text normalization~\cite{chrupala2014normalizing}, part-of-speech tagging~\cite{ling2015finding} and parsing~\cite{ballesteros2015improved}.
Another family of models are convolutional neural networks trained on characters, which were applied to part-of-speech tagging~\cite{santos2014learning}, sentiment analysis~\cite{santos2014deep}, text classification~\cite{zhang2015character} and language modeling~\cite{kim2016character}.
\newcite{sperr2013letter} introduced a language model based on restricted Boltzmann machines, in which words are encoded as a set of character $n$-grams.
Finally, recent works in machine translation have proposed using subword units to obtain representations of rare words~\cite{sennrich2016neural,luong2016hybrid}.

%% file: model.tex
\section{Model}
In this section, we propose our model to learn word representations while taking into account morphology.
We model morphology by considering subword units, and representing words by a sum of its character $n$-grams.
We will begin by presenting the general framework that we use to train word vectors, then present our subword model and eventually describe how we handle the dictionary of character $n$-grams.

\subsection{General model}
We start by briefly reviewing the continuous skipgram model introduced by \newcite{mikolov2013distributed}, from which our model is derived.
Given a word vocabulary of size $W$, where a word is identified by its index $w~\in~\{1, ..., W\}$, the goal is to learn a vectorial representation for each word $w$.
Inspired by the distributional hypothesis~\cite{harris1954distributional}, word representations are trained to \emph{predict well} words that appear in its context.
More formally, given a large training corpus represented as a sequence of words $w_1, ..., w_T$, the objective of the skipgram model is to maximize the following log-likelihood:
\begin{equation*}
  \sum_{t=1}^T \ \sum_{c \in \mathcal{C}_t} \ \log p(w_c \ | \ w_t),
\end{equation*}
where the context $\mathcal{C}_t$ is the set of indices of words surrounding word $w_t$.
The probability of observing a context word $w_c$ given $w_t$ will be parameterized using the aforementioned word vectors.
For now, let us consider that we are given a scoring function~$s$ which maps pairs of (word, context) to scores in~$\mathbb{R}$.
One possible choice to define the probability of a context word is the softmax:
\begin{equation*}
p(w_c \ | \ w_t) = \frac{e^{s(w_t,\ w_c)}}{\sum_{j=1}^W e^{s(w_t,\ j)}}.
\end{equation*}
However, such a model is not adapted to our case as it implies that, given a word $w_t$, we only predict one context word $w_c$.

The problem of predicting context words can instead be framed as a set of independent binary classification tasks.
Then the goal is to independently predict the presence (or absence) of context words.
For the word at position $t$ we consider all context words as positive examples and sample negatives at random from the dictionary.
For a chosen context position $c$, using the binary logistic loss, we obtain the following negative log-likelihood:
\begin{equation*}
  \log \left(1 + e^{-s(w_t,\ w_c)} \right) + \sum_{n \in \mathcal{N}_{t, c}} \log \left(1 + e^{s(w_t,\ n)}\right),
\end{equation*}
where $\mathcal{N}_{t,c}$ is a set of negative examples sampled from the vocabulary.
By denoting the logistic loss function $\ell: x \mapsto \log(1 + e^{-x})$, we can re-write the objective as:
\begin{equation*}
\sum_{t=1}^{T}  \left [ \sum_{c \in \mathcal{C}_t} \ell(s(w_t,\ w_c)) + \sum_{n \in \mathcal{N}_{t,c}} \ell(-s(w_t,\ n)) \right ].
\end{equation*}
A natural parameterization for the scoring function $s$ between a word $w_t$ and a context word $w_c$ is to use word vectors.
Let us define for each word $w$ in the vocabulary two vectors $u_w$ and $v_w$ in $\mathbb{R}^d$.
These two vectors are sometimes referred to as \emph{input} and \emph{output} vectors in the literature.
In particular, we have vectors $\mathbf{u}_{w_t}$ and $\mathbf{v}_{w_c}$, corresponding, respectively, to words $w_t$ and $w_c$.
Then the score can be computed as the scalar product between word and context vectors as $s(w_t, w_c) = \mathbf{u}_{w_t}^{\top} \mathbf{v}_{w_c}$.
The model described in this section is the skipgram model with negative sampling, introduced by \newcite{mikolov2013distributed}.

\subsection{Subword model}
\label{sec:model-ngrams}

By using a distinct vector representation for each word, the skipgram model ignores the internal structure of words.
In this section, we propose a different scoring function $s$, in order to take into account this information.

Each word $w$ is represented as a bag of character $n$-gram.
We add special boundary symbols \texttt{<} and \texttt{>} at the beginning and end of words, allowing to distinguish prefixes and suffixes from other character sequences.
We also include the word $w$ itself in the set of its $n$-grams, to learn a representation for each word (in addition to character $n$-grams).
Taking the word \emph{where} and $n=3$ as an example, it will be represented by the character $n$-grams:
\begin{center}
\texttt{<wh, whe, her, ere, re>}
\end{center}
and the special sequence
\begin{center}
\texttt{<where>}.
\end{center}
Note that the sequence \texttt{<her>}, corresponding to the word \emph{her} is different from the tri-gram \texttt{her} from the word \emph{where}.
In practice, we extract all the $n$-grams for $n$ greater or equal to 3 and smaller or equal to $6$.
This is a very simple approach, and different sets of $n$-grams could be considered, for example taking all prefixes and suffixes.

Suppose that you are given a dictionary of $n$-grams of size $G$.
Given a word $w$, let us denote by $\mathcal{G}_w \subset \{1, \dots, G \}$ the set of $n$-grams appearing in $w$.
We associate a vector representation $\mathbf{z}_g$ to each $n$-gram $g$.
We represent a word by the sum of the vector representations of its $n$-grams.
We thus obtain the scoring function:
\begin{equation*}
s(w, c) = \sum_{g \in \mathcal{G}_w} \mathbf{z}_g^\top \mathbf{v}_c.
\end{equation*}
This simple model allows sharing the representations across words, thus allowing to learn reliable representation for rare words.

In order to bound the memory requirements of our model, we use a hashing function that maps $n$-grams to integers in 1 to $K$.
We hash character sequences using the Fowler-Noll-Vo hashing function (specifically the \texttt{FNV-1a} variant).\footnote{\smaller\relax\url{http://www.isthe.com/chongo/tech/comp/fnv}}
We set $K = 2.10^6$ below.
Ultimately, a word is represented by its index in the word dictionary and the set of hashed $n$-grams it contains.

%% file: experiments.tex
%%%%%%%%%%%%%%%%%%%
%%% EXPERIMENTS %%%
%%%%%%%%%%%%%%%%%%%

\section{Experimental setup}

\subsection{Baseline}
In most experiments (except in Sec.~\ref{sec:sota}), we compare our model to the C implementation of the \texttt{skipgram} and \texttt{cbow} models from the \texttt{word2vec}\footnote{\smaller\relax\url{https://code.google.com/archive/p/word2vec}} package.

\subsection{Optimization}
We solve our optimization problem by performing stochastic gradient descent on the negative log likelihood presented before.
As in the baseline \texttt{skipgram} model, we use a linear decay of the step size.
Given a training set containing $T$ words and a number of passes over the data equal to $P$, the step size at time $t$ is equal to $\gamma_0 (1 - \frac{t}{TP})$, where $\gamma_0$ is a fixed parameter.
%Suppose that we want to make $P$ passes a dataset composed of $T$ words, the step size at time $t$ is equal to $\gamma_0 (1 - \frac{t}{T})$, where $\gamma_0$ is a fixed parameter.
We carry out the optimization in parallel, by resorting to Hogwild~\cite{recht2011hogwild}. 
All threads share parameters and update vectors in an asynchronous manner.

\subsection{Implementation details}
For both our model and the baseline experiments, we use the following parameters: the word vectors have dimension $300$.
For each positive example, we sample $5$ negatives at random, with probability proportional to the square root of the uni-gram frequency.
We use a context window of size $c$, and uniformly sample the size $c$ between $1$ and $5$.
In order to subsample the most frequent words, we use a rejection threshold of $10^{-4}$ (for more details, see \cite{mikolov2013distributed}).
When building the word dictionary, we keep the words that appear at least $5$ times in the training set.
The step size $\gamma_0$ is set to $0.025$ for the \texttt{skipgram} baseline and to $0.05$ for both our model and the \texttt{cbow} baseline.
These are the default values in the \texttt{word2vec} package and work well for our model too.

Using this setting on English data, our model with character $n$-grams is approximately $1.5 \times$ slower to train than the \texttt{skipgram} baseline.
Indeed, we process $105$k words/second/thread versus $145$k words/second/thread for the baseline.
Our model is implemented in C++, and is publicly available.\footnote{\smaller\relax\url{https://github.com/facebookresearch/fastText}}

\subsection{Datasets}
Except for the comparison to previous work~(Sec.~\ref{sec:sota}), we train our models on Wikipedia data.\footnote{\smaller\relax\url{https://dumps.wikimedia.org}}
We downloaded Wikipedia dumps in nine languages: Arabic, Czech, German, English, Spanish, French, Italian, Romanian and Russian.
We normalize the raw Wikipedia data using Matt Mahoney's pre-processing perl script.\footnote{\smaller\relax\url{http://mattmahoney.net/dc/textdata}}
All the datasets are shuffled, and we train our models by doing five passes over them.

\section{Results}

We evaluate our model in five experiments: an evaluation of word similarity and word analogies, a comparison to state-of-the-art methods, an analysis of the effect of the size of training data and of the size of character $n$-grams that we consider.
We will describe these experiments in detail in the following sections.

%%%%%%%%%%%%%%%%%%
%%% SIMILARITY %%%
%%%%%%%%%%%%%%%%%%

\subsection{Human similarity judgement}
\label{sec:wordsim}

We first evaluate the quality of our representations on the task of word similarity / relatedness.
We do so by computing Spearman's rank correlation coefficient~\cite{spearman04proof} between human judgement and the cosine similarity between the vector representations.
For German, we compare the different models on three datasets: \textsc{Gur65}, \textsc{Gur350} and \textsc{ZG222}~\cite{gurevych2005using,zesch2006automatically}.
For English, we use the \textsc{WS353} dataset introduced by \newcite{finkelstein2001placing} and the rare word dataset (\textsc{RW}), introduced by \newcite{luong2013better}.
We evaluate the French word vectors on the translated dataset \textsc{RG65}~\cite{joubarne2011comparison}.
Spanish, Arabic and Romanian word vectors are evaluated using the datasets described in~\cite{hassan2009cross}.
Russian word vectors are evaluated using the \textsc{HJ} dataset introduced by \newcite{panchenko2016human}.

\begin{table}[t]
  \centering
  \begin{tabular}{@{}rrcccc@{}}
    \toprule
    && \texttt{sg} & \texttt{cbow} & \texttt{sisg-} & \texttt{sisg} \\
    \midrule
                     \textsc{Ar}  & \textsc{WS353}  & 51 & 52 & 54 & \textbf{55} \\
    \midrule
    \multirow{ 3}{*}{\textsc{De}} & \textsc{Gur350} & 61 & 62 & 64 & \textbf{70} \\
                                  & \textsc{Gur65}  & 78 & 78 & \textbf{81} & \textbf{81} \\
                                  & \textsc{ZG222}  & 35 & 38 & 41 & \textbf{44} \\
    \midrule
    \multirow{ 2}{*}{\textsc{En}} & \textsc{RW}     & 43 & 43 & 46 & \textbf{47} \\
                                  & \textsc{WS353}  & 72 & \textbf{73} & 71 & 71 \\
    \midrule
    \textsc{Es}                   & \textsc{WS353}  & 57 & 58 & 58 & \textbf{59} \\
    \midrule
    \textsc{Fr}                   & \textsc{RG65}   & 70 & 69 & \textbf{75} & \textbf{75} \\
    \midrule
    \textsc{Ro}                   & \textsc{WS353}  & 48 & 52 & 51 & \textbf{54} \\
    \midrule
    \textsc{Ru}                   & \textsc{HJ}     & 59 & 60 & 60 & \textbf{66} \\
    \bottomrule
  \end{tabular}
  \caption{
    Correlation between human judgement and similarity scores on word similarity datasets.
    We train both our model and the \texttt{word2vec} baseline on normalized Wikipedia dumps.
    Evaluation datasets contain words that are not part of the training set, so we represent them using null vectors (\texttt{sisg-}).
    With our model, we also compute vectors for unseen words by summing the $n$-gram vectors (\texttt{sisg}).
  }
  \label{tab:wordsim}
\end{table}

We report results for our method and baselines for all datasets in Table~\ref{tab:wordsim}. 
Some words from these datasets do not appear in our training data, and thus, we cannot obtain word representation for these words using the \texttt{cbow} and \texttt{skipgram} baselines.
In order to provide comparable results, we propose by default to use null vectors for these words.
Since our model exploits subword information, we can also compute valid representations for out-of-vocabulary words.
We do so by taking the sum of its $n$-gram vectors.
When OOV words are represented using null vectors we refer to our method as \texttt{sisg-} and \texttt{sisg} otherwise (Subword Information Skip Gram).

First, by looking at Table~\ref{tab:wordsim}, we notice that the proposed model (\texttt{sisg}), which uses subword information, outperforms the baselines on all datasets except the English \textsc{WS353} dataset.
Moreover, computing vectors for out-of-vocabulary words (\texttt{sisg}) is always at least as good as not doing so (\texttt{sisg-}).
This proves the advantage of using subword information in the form of character $n$-grams.

Second, we observe that the effect of using character $n$-grams is more important for Arabic, German and Russian than for English, French or Spanish.
German and Russian exhibit grammatical declensions with four cases for German and six for Russian.
Also, many German words are compound words; for instance the nominal phrase ``table tennis'' is written in a single word as ``Tischtennis''.
By exploiting the character-level similarities between ``Tischtennis'' and ``Tennis'', our model does not represent the two words as completely different words.

Finally, we observe that on the English Rare Words dataset (\textsc{RW}), our approach outperforms the baselines while it does not on the English \textsc{WS353} dataset.
This is due to the fact that words in the English \textsc{WS353} dataset are common words for which good vectors can be obtained without exploiting subword information.
When evaluating on less frequent words, we see that using similarities at the character level between words can help learning good word vectors.

%%%%%%%%%%%%%%%%%
%%% ANALOGIES %%%
%%%%%%%%%%%%%%%%%

\subsection{Word analogy tasks}

\begin{table}[t]
  \centering
  \begin{tabular}{rrccc}
    \toprule
    & & \texttt{sg} & \texttt{cbow} & \texttt{sisg} \\
    \midrule
    \multirow{ 2}{*}{\textsc{Cs}} & Semantic  & 25.7 & 27.6 & 27.5 \\
                                  & Syntactic & 52.8 & 55.0 & 77.8 \\
    \midrule
    \multirow{ 2}{*}{\textsc{De}} & Semantic  & 66.5 & 66.8 & 62.3 \\
                                  & Syntactic & 44.5 & 45.0 & 56.4 \\
    \midrule
    \multirow{ 2}{*}{\textsc{En}} & Semantic  & 78.5 & 78.2 & 77.8 \\
                                  & Syntactic & 70.1 & 69.9 & 74.9 \\
    \midrule
    \multirow{ 2}{*}{\textsc{It}} & Semantic  & 52.3 & 54.7 & 52.3 \\
                                  & Syntactic & 51.5 & 51.8 & 62.7 \\
    \bottomrule
  \end{tabular}
  \caption{
    Accuracy of our model and baselines on word analogy tasks for Czech, German, English and Italian.
    We report results for semantic and syntactic analogies separately.
  %  The proposed approach (\texttt{our}) significantly outperforms baselines on syntactic analogies, while keeping good semantic performance.
  }
  \label{tab:wordanalogy}
\end{table}

We now evaluate our approach on word analogy questions, of the form $A$ is to $B$ as $C$ is to $D$, where $D$ must be predicted by the models.
We use the datasets introduced by \newcite{mikolov2013efficient} for English, by \newcite{svoboda2016new} for Czech, by \newcite{koper2015multilingual} for German and by \newcite{berardi2015word} for Italian.
Some questions contain words that do not appear in our training corpus, and we thus excluded these questions from the evaluation.

\begin{table*}[t]
  \centering
  \begin{tabular}{rcccccccccc}
    \toprule
    && \multicolumn{2}{c}{\textsc{De}} && \multicolumn{2}{c}{\textsc{En}} && \multicolumn{1}{c}{\textsc{Es}} && \multicolumn{1}{c}{\textsc{Fr}}  \\
    \cmidrule{3-4} \cmidrule{6-7} \cmidrule{9-9} \cmidrule{11-11}
    && \textsc{Gur350} & \textsc{ZG222} && WS353 & RW && WS353 && RG65  \\
    \midrule
    \newcite{luong2013better}         && -  & -  && 64 & 34 && -  && -  \\
    \newcite{qiu2014colearning}       && -  & -  && 65 & 33 && -  && -  \\
    \newcite{soricut2015unsupervised} && 64 & 22 && 71 & 42 && 47 && 67 \\
    \texttt{sisg}                    && 73 & 43 && 73 & 48 && 54 && 69 \\
    \midrule
    \newcite{botha2014compositional}  && 56 & 25 && 39 & 30 && 28 && 45 \\
    \texttt{sisg}                    && 66 & 34 && 54 & 41 && 49 && 52 \\
    \bottomrule
  \end{tabular}
  \caption{
    Spearman's rank correlation coefficient between human judgement and model scores for different methods using morphology to learn word representations.
    We keep all the word pairs of the evaluation set and obtain representations for out-of-vocabulary words with our model by summing the vectors of character $n$-grams.
    Our model was trained on the same datasets as the methods we are comparing to (hence the two lines of results for our approach).
  }
  \label{tab:comparison}
\end{table*}

We report accuracy for the different models in Table~\ref{tab:wordanalogy}.
We observe that morphological information significantly improves the syntactic tasks; our approach outperforms the baselines.
In contrast, it does not help for semantic questions, and even degrades the performance for German and Italian.
Note that this is tightly related to the choice of the length of character $n$-grams that we consider.
We show in Sec.~\ref{sec:ngram-size} that when the size of the $n$-grams is chosen optimally, the semantic analogies degrade less.
Another interesting observation is that, as expected, the improvement over the baselines is more important for morphologically rich languages, such as Czech and German.

%%%%%%%%%%%%%%%%%%%%%%%%%%
%%% COMPARISON TO SOTA %%%
%%%%%%%%%%%%%%%%%%%%%%%%%%

\subsection{Comparison with morphological representations}
\label{sec:sota}

We also compare our approach to previous work on word vectors incorporating subword information on word similarity tasks.
The methods used are: the recursive neural network of \newcite{luong2013better}, the morpheme \texttt{cbow} of \newcite{qiu2014colearning} and the morphological transformations of \newcite{soricut2015unsupervised}.
In order to make the results comparable, we trained our model on the same datasets as the methods we are comparing to: the English Wikipedia data released by \newcite{shaoul2010westbury}, and the news crawl data from the 2013 WMT shared task for German, Spanish and French.
We also compare our approach to the log-bilinear language model introduced by \newcite{botha2014compositional}, which was trained on the Europarl and news commentary corpora.
Again, we trained our model on the same data to make the results comparable.
Using our model, we obtain representations of out-of-vocabulary words by summing the representations of character $n$-grams.
We report results in Table~\ref{tab:comparison}.
We observe that our simple approach performs well relative to techniques based on subword information obtained from morphological segmentors.
We also observe that our approach outperforms the \newcite{soricut2015unsupervised} method, which is based on prefix and suffix analysis.
The large improvement for German is due to the fact that their approach does not model noun compounding, contrary to ours.

%%%%%%%%%%%%%%%%%
%%% DATA SIZE %%%
%%%%%%%%%%%%%%%%%

\subsection{Effect of the size of the training data}
\label{sec:exp-data-size}

\begin{figure*}[t]
    \centering
    \begin{subfigure}[b]{0.43\textwidth}
        \includegraphics[width=\textwidth]{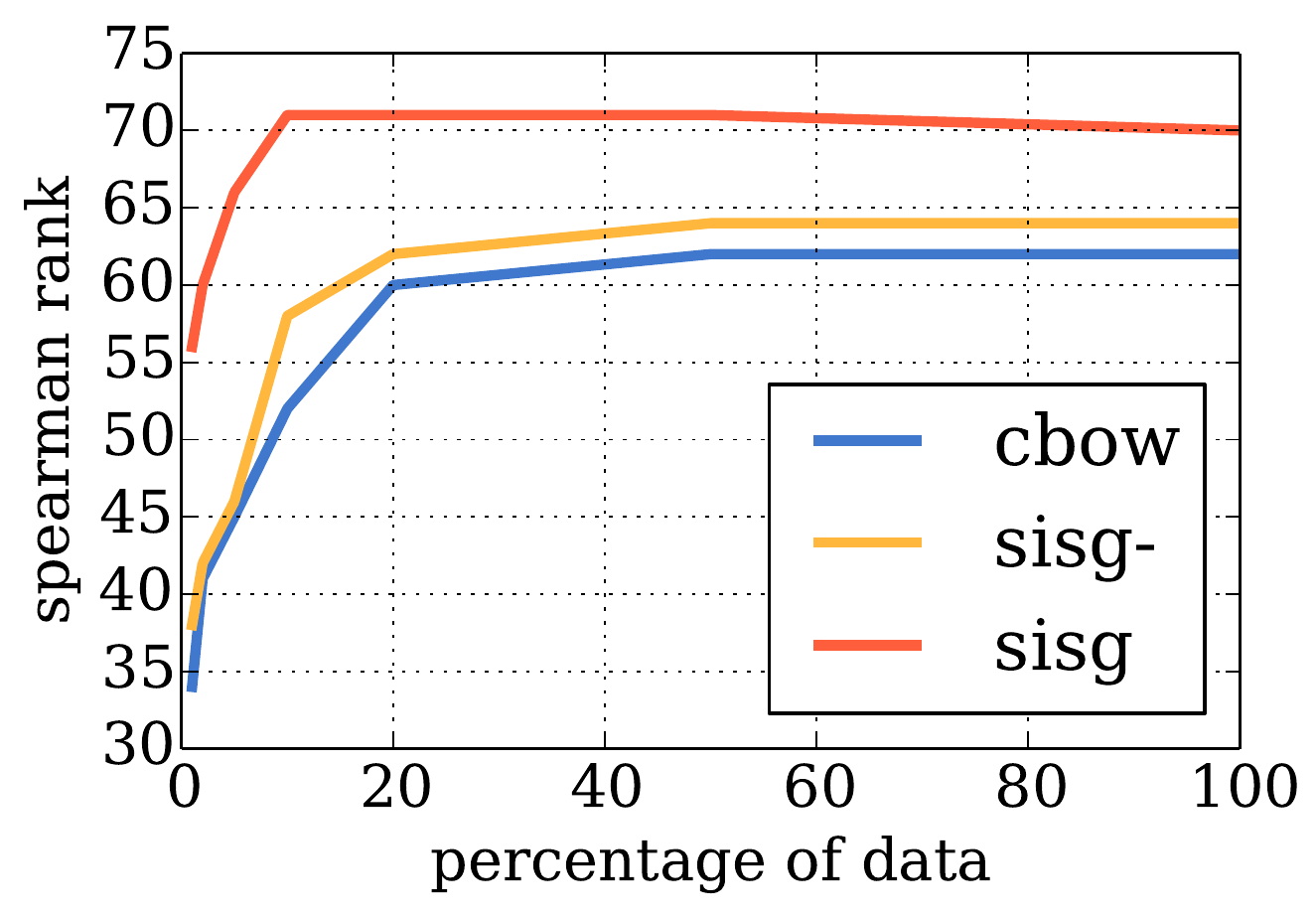}
        \caption{\textsc{De-Gur350}}
    \end{subfigure}
    \qquad
    \begin{subfigure}[b]{0.43\textwidth}
        \includegraphics[width=\textwidth]{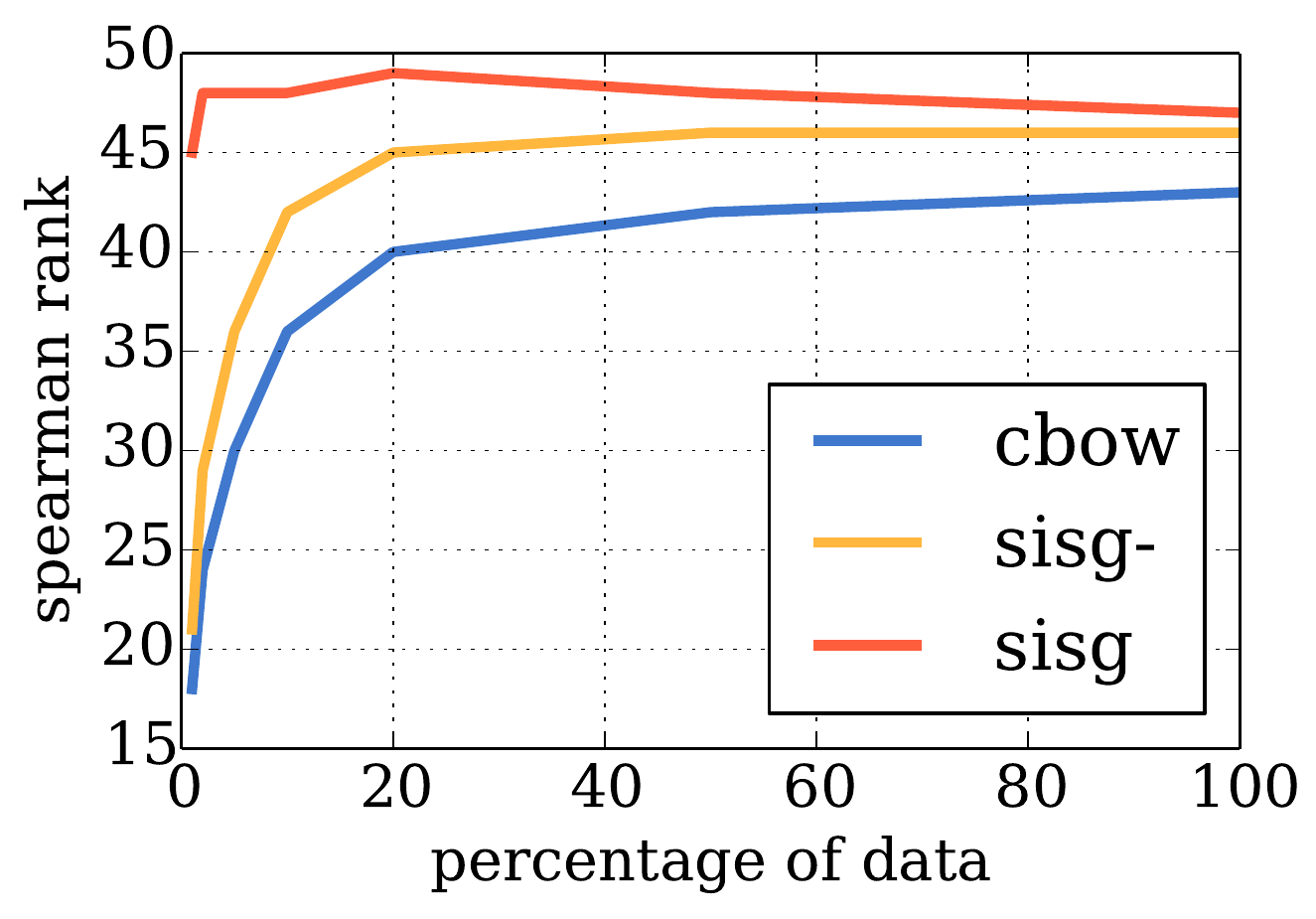}
        \caption{\textsc{En-RW}}
    \end{subfigure}
    \caption{
        Influence of size of the training data on performance.
        We compute word vectors following the proposed model using datasets of increasing size.
        In this experiment, we train models on a fraction of the full Wikipedia dump.
    }
    \label{fig:gulllss}
\end{figure*}

Since we exploit character-level similarities between words, we are able to better model infrequent words.
Therefore, we should also be more robust to the size of the training data that we use.
In order to assess that, we propose to evaluate the performance of our word vectors on the similarity task as a function of the training data size.
To this end, we train our model and the \texttt{cbow} baseline on portions of Wikipedia of increasing size.
We use the Wikipedia corpus described above and isolate the first $1$, $2$, $5$, $10$, $20$, and $50$ percent of the data.
Since we don't reshuffle the dataset, they are all subsets of each other.
We report results in Fig.~\ref{fig:gulllss}.

As in the experiment presented in Sec.~\ref{sec:wordsim}, not all words from the evaluation set are present in the Wikipedia data.
Again, by default, we use a null vector for these words (\texttt{sisg-}) or compute a vector by summing the $n$-gram representations (\texttt{sisg}).
The out-of-vocabulary rate is growing as the dataset shrinks, and therefore the performance of \texttt{sisg-} and \texttt{cbow} necessarily degrades.
However, the proposed model (\texttt{sisg}) assigns non-trivial vectors to previously unseen words.

First, we notice that for all datasets, and all sizes, the proposed approach (\texttt{sisg}) performs better than the baseline.
However, the performance of the baseline \texttt{cbow} model gets better as more and more data is available.
Our model, on the other hand, seems to quickly saturate and adding more data does not always lead to improved results.

Second, and most importantly, we notice that the proposed approach provides very good word vectors even when using very small training datasets.
For instance, on the German \textsc{Gur350} dataset, our model (\texttt{sisg}) trained on $5\%$ of the data achieves better performance ($66$) than the \texttt{cbow} baseline trained on the full dataset ($62$).
On the other hand, on the English \textsc{RW} dataset, using $1\%$ of the Wikipedia corpus we achieve a correlation coefficient of $45$ which is better than the performance of \texttt{cbow} trained on the full dataset ($43$).
This has a very important practical implication: well performing word vectors can be computed on datasets of a restricted size and still work well on previously unseen words. 
In general, when using vectorial word representations in specific applications, it is recommended to retrain the model on textual data relevant for the application.
However, this kind of relevant task-specific data is often very scarce and learning from a reduced amount of training data is a great advantage.

%%%%%%%%%%%%%%
%%% NGRAMS %%%
%%%%%%%%%%%%%%

\subsection{Effect of the size of $n$-grams}
\label{sec:ngram-size}

The proposed model relies on the use of character $n$-grams to represent words as vectors.
As mentioned in Sec.~\ref{sec:model-ngrams}, we decided to use $n$-grams ranging from $3$ to $6$ characters.
This choice was arbitrary, motivated by the fact that $n$-grams of these lengths will cover a wide range of information.
They would include short suffixes (corresponding to conjugations and declensions for instance) as well as longer roots.
In this experiment, we empirically check for the influence of the range of $n$-grams that we use on performance.
We report our results in Table~\ref{tab:nsize} for English and German on word similarity and analogy datasets.

\begin{table*}[t]
  \centering
  \begin{subtable}[b]{0.31\textwidth}
    \begin{tabular}{c c c c c c}
      \toprule
      & 2 & 3 & 4 & 5 & 6 \\
      \midrule
      2 & 57 & 64 & 67 & 69 & 69  \\
      3 &    & 65 & 68 & 70 & 70  \\
      4 &    &    & 70 & 70 & \textbf{71}  \\
      5 &    &    &    & 69 & \textbf{71}  \\
      6 &    &    &    &    & 70  \\
      \bottomrule
    \end{tabular}
    \caption{\textsc{De-Gur350}}
  \end{subtable}
  \begin{subtable}[b]{0.31\textwidth}
    \begin{tabular}{c c c c c c}
      \toprule
      & 2 & 3 & 4 & 5 & 6 \\
      \midrule
      2 & 59 & 55 & 56 & 59 & 60 \\
      3 &    & 60 & 58 & 60 & 62 \\
      4 &    &    & 62 & 62 & 63 \\ 
      5 &    &    &    & 64 & 64 \\
      6 &    &    &    &    & \textbf{65} \\
      \bottomrule
    \end{tabular}
    \caption{\textsc{De} Semantic}
  \end{subtable}
  \begin{subtable}[b]{0.31\textwidth}
    \centering
    \begin{tabular}{c c c c c c}
      \toprule
      & 2 & 3 & 4 & 5 & 6 \\
      \midrule
      2 & 45 & 50 & 53 & 54 & 55 \\ 
      3 &    & 51 & 55 & 55 & \textbf{56} \\
      4 &    &    & 54 & \textbf{56} & \textbf{56} \\
      5 &    &    &    & \textbf{56} & \textbf{56} \\ 
      6 &    &    &    &    & 54 \\
      \bottomrule
    \end{tabular}
    \caption{\textsc{De} Syntactic}
  \end{subtable}

  \vspace{1em}

  \begin{subtable}[b]{0.31\textwidth}
    \centering
    \begin{tabular}{c c c c c c}
      \toprule
          & 2 & 3 & 4 & 5 & 6 \\
      \midrule
      2 & 41 & 42 & 46 & 47 & \textbf{48}  \\
      3 &    & 44 & 46 & \textbf{48} & \textbf{48}  \\
      4 &    &    & 47 & \textbf{48} & \textbf{48}  \\
      5 &    &    &    & \textbf{48} & \textbf{48}  \\
      6 &    &    &    &    & \textbf{48}  \\
      \bottomrule
    \end{tabular}
    \caption{\textsc{En-RW}}
  \end{subtable}
  \begin{subtable}[b]{0.31\textwidth}
    \centering
    \begin{tabular}{c c c c c c}
      \toprule
      & 2 & 3 & 4 & 5 & 6 \\
      \midrule
      2 & 78 & 76 & 75 & 76 & 76 \\
      3 &    & 78 & 77 & 78 & 77 \\
      4 &    &    & 79 & 79 & 79 \\
      5 &    &    &    & \textbf{80} & 79 \\
      6 &    &    &    &    & \textbf{80} \\
      \bottomrule
    \end{tabular}
    \caption{\textsc{En} Semantic}
  \end{subtable}
  \begin{subtable}[b]{0.31\textwidth}
    \centering
    \begin{tabular}{c c c c c c}
      \toprule
      & 2 & 3 & 4 & 5 & 6 \\
      \midrule
      2 & 70 & 71 & 73 & 74 & 73 \\
      3 &    & 72 & 74 & \textbf{75} & 74 \\
      4 &    &    & 74 & \textbf{75} & \textbf{75} \\
      5 &    &    &    & 74 & 74 \\
      6 &    &    &    &    & 72 \\
      \bottomrule
    \end{tabular}
    \caption{\textsc{En} Syntactic}
  \end{subtable}
  \caption{
    Study of the effect of sizes of $n$-grams considered on performance.
    We compute word vectors by using character $n$-grams with $n$ in $\{i, \dots, j\}$ and report performance for various values of $i$ and $j$.
    We evaluate this effect on German and English, and represent out-of-vocabulary words using subword information.
  }
  \label{tab:nsize}
\end{table*}

We observe that for both English and German, our arbitrary choice of $3$-$6$ was a reasonable decision, as it provides satisfactory performance across languages.
The optimal choice of length ranges depends on the considered task and language and should be tuned appropriately.
However, due to the scarcity of test data, we did not implement any proper validation procedure to automatically select the best parameters.
Nonetheless, taking a large range such as $3-6$ provides a reasonable amount of subword information.

This experiment also shows that it is important to include long $n$-grams, as columns corresponding to $n \leq 5$ and $n \leq 6$ work best.
This is especially true for German, as many nouns are compounds made up from several units that can only be captured by longer character sequences.
On analogy tasks, we observe that using larger $n$-grams helps for semantic analogies.
However, results are always improved by taking $n \geq 3$ rather than $n \geq 2$, which shows that character $2$-grams are not informative for that task.
As described in Sec.~\ref{sec:model-ngrams}, before computing character $n$-grams, we prepend and append special positional characters to represent the beginning and end of word.
Therefore, $2$-grams will not be enough to properly capture suffixes that correspond to conjugations or declensions, since they are composed of a single proper character and a positional one.

\subsection{Language modeling}

In this section, we describe an evaluation of the word vectors obtained with our method on a language modeling task.
We evaluate our language model on five languages (\textsc{Cs}, \textsc{De}, \textsc{Es}, \textsc{Fr}, \textsc{Ru}) using the datasets introduced by \newcite{botha2014compositional}.
Each dataset contains roughly one million training tokens, and we use the same preprocessing and data splits as \newcite{botha2014compositional}.

Our model is a recurrent neural network with $650$ LSTM units, regularized with dropout (with probability of $0.5$) and weight decay (regularization parameter of $10^{-5}$).
We learn the parameters using the Adagrad algorithm with a learning rate of $0.1$, clipping the gradients which have a norm larger than~$1.0$.
We initialize the weight of the network in the range $[-0.05, 0.05]$, and use a batch size of $20$.
Two baselines are considered: we compare our approach to the log-bilinear language model of \newcite{botha2014compositional} and the character aware language model of \newcite{kim2016character}.
We trained word vectors with character $n$-grams on the training set of the language modeling task and use them to initialize the lookup table of our language model.
We report the test perplexity of our model without using pre-trained word vectors (\texttt{LSTM}), with word vectors pre-trained without subword information (\texttt{sg}) and with our vectors (\texttt{sisg}).
The results are presented in Table~\ref{tab:lm}.

\begin{table}[t]
  \centering
  \begin{tabular}{rccccc}
    \toprule
                      & \textsc{Cs}  & \textsc{De}  & \textsc{Es}  & \textsc{Fr}  & \textsc{Ru}  \\
    \midrule
    Vocab. size       & 46k & 37k & 27k & 25k & 63k \\
    \midrule
    \textsc{CLBL}     & 465 & 296 & 200 & 225 & 304 \\
    \textsc{CANLM}    & 371 & 239 & 165 & 184 & 261 \\
    \midrule
    \texttt{LSTM}     & 366 & 222 & 157 & 173 & 262 \\
    \texttt{sg}       & 339 & 216 & 150 & 162 & 237 \\
    \texttt{sisg}    & \textbf{312} & \textbf{206} & \textbf{145} & \textbf{159} & \textbf{206} \\
    \bottomrule
  \end{tabular}
  \caption{
    Test perplexity on the language modeling task, for 5 different languages.
    We compare to two state of the art approaches: \textsc{CLBL} refers to the work of \protect\newcite{botha2014compositional} and \textsc{CANLM} refers to the work of \protect\newcite{kim2016character}.
  }
  \label{tab:lm}
\end{table}

\begin{table}[t]
  \centering
  \small
  \begin{tabular}{rrrrr}
\toprule 
&                word &         \multicolumn{3}{c}{$n$-grams}        \\
\midrule
   &       autofahrer &    fahr &  fahrer &    auto \\ 
   &    freundeskreis &   kreis &  kreis> &  <freun \\ 
\textsc{De} &        grundwort &    wort &   wort> &   grund \\ 
   &     sprachschule &   schul &  hschul &  sprach \\ 
   &       tageslicht &   licht &    gesl &   tages \\ 
\midrule
   &          anarchy &     chy &   <anar &  narchy \\ 
   &         monarchy &  monarc &     chy &  <monar \\ 
   &         kindness &   ness> &    ness &    kind \\ 
   &       politeness &  polite &   ness> &  eness> \\ 
\textsc{En} &          unlucky &     <un &    cky> &  nlucky \\ 
   &         lifetime &    life &   <life &    time \\ 
   &         starfish &    fish &   fish> &    star \\ 
   &        submarine &  marine &     sub &   marin \\ 
   &        transform &   trans &  <trans &    form \\ 
\midrule
   &         finirais &    ais> &     nir &    fini \\ 
\textsc{Fr} &        finissent &    ent> &  finiss &  <finis \\ 
   &       finissions &   ions> &  finiss &  sions> \\ 
\bottomrule
  \end{tabular}
  \caption{
		Illustration of most important character $n$-grams for selected words in three languages.
    For each word, we show the $n$-grams that, when removed, result in the most different representation.
  }
  \label{tab:morphemes}
\end{table}

We observe that initializing the lookup table of the language model with pre-trained word representations improves the test perplexity over the baseline LSTM.
The most important observation is that using word representations trained with subword information outperforms the plain skipgram model.
We observe that this improvement is most significant for morphologically rich Slavic languages such as Czech (8\% reduction of perplexity over \texttt{sg}) and Russian (13\% reduction).
The improvement is less significant for Roman languages such as Spanish (3\% reduction) or French (2\% reduction).
This shows the importance of subword information on the language modeling task and exhibits the usefulness of the vectors that we propose for morphologically rich languages.

%%%%%%%%%%%%%%%%%%%%%%%%%%%
%%% QUALITATIVE RESULTS %%%
%%%%%%%%%%%%%%%%%%%%%%%%%%%

\section{Qualitative analysis}

\subsection{Nearest neighbors.}

We report sample qualitative results in Table~\ref{tab:nn}.
For selected words, we show nearest neighbors according to cosine similarity for vectors trained using the proposed approach and for the \texttt{skipgram} baseline.
As expected, the nearest neighbors for complex, technical and infrequent words using our approach are better than the ones obtained using the baseline model.

\subsection{Character $n$-grams and morphemes}
We want to qualitatively evaluate whether or not the most important $n$-grams in a word correspond to morphemes.
To this end, we take a word vector that we construct as the sum of $n$-grams.
As described in Sec.~\ref{sec:model-ngrams}, each word $w$ is represented as the sum of its $n$-grams: $u_w = \sum_{g \in \mathcal{G}_w} z_g$.
For each $n$-gram~$g$, we propose to compute the restricted representation~$u_{w \backslash g}$ obtained by omitting $g$: 
\begin{equation}
  u_{w \backslash g} = \sum_{g' \in \mathcal{G} - \{g\}} z_{g'}.\nonumber
\end{equation}
We then rank $n$-grams by increasing value of cosine between $u_w$ and $u_{w \backslash g}$.	
We show ranked $n$-grams for selected words in three languages in Table~\ref{tab:morphemes}. 

For German, which has a lot of compound nouns, we observe that the most important $n$-grams correspond to valid morphemes.
Good examples include \emph{Autofahrer} (car driver)  whose most important $n$-grams are \emph{Auto} (car) and \emph{Fahrer} (driver).
We also observe the separation of compound nouns into morphemes in English, with words such as \emph{lifetime} or \emph{starfish}.
However, for English, we also observe that $n$-grams can correspond to affixes in words such as \emph{kindness} or \emph{unlucky}.
Interestingly, for French we observe the inflections of verbs with endings such as \emph{ais>}, \emph{ent>} or \emph{ions>}.

\subsection{Word similarity for OOV words}
\begin{figure*}[p]
  \begin{minipage}[t]{.45\textwidth}
    \vspace{0pt}
    \includegraphics[width=\linewidth]{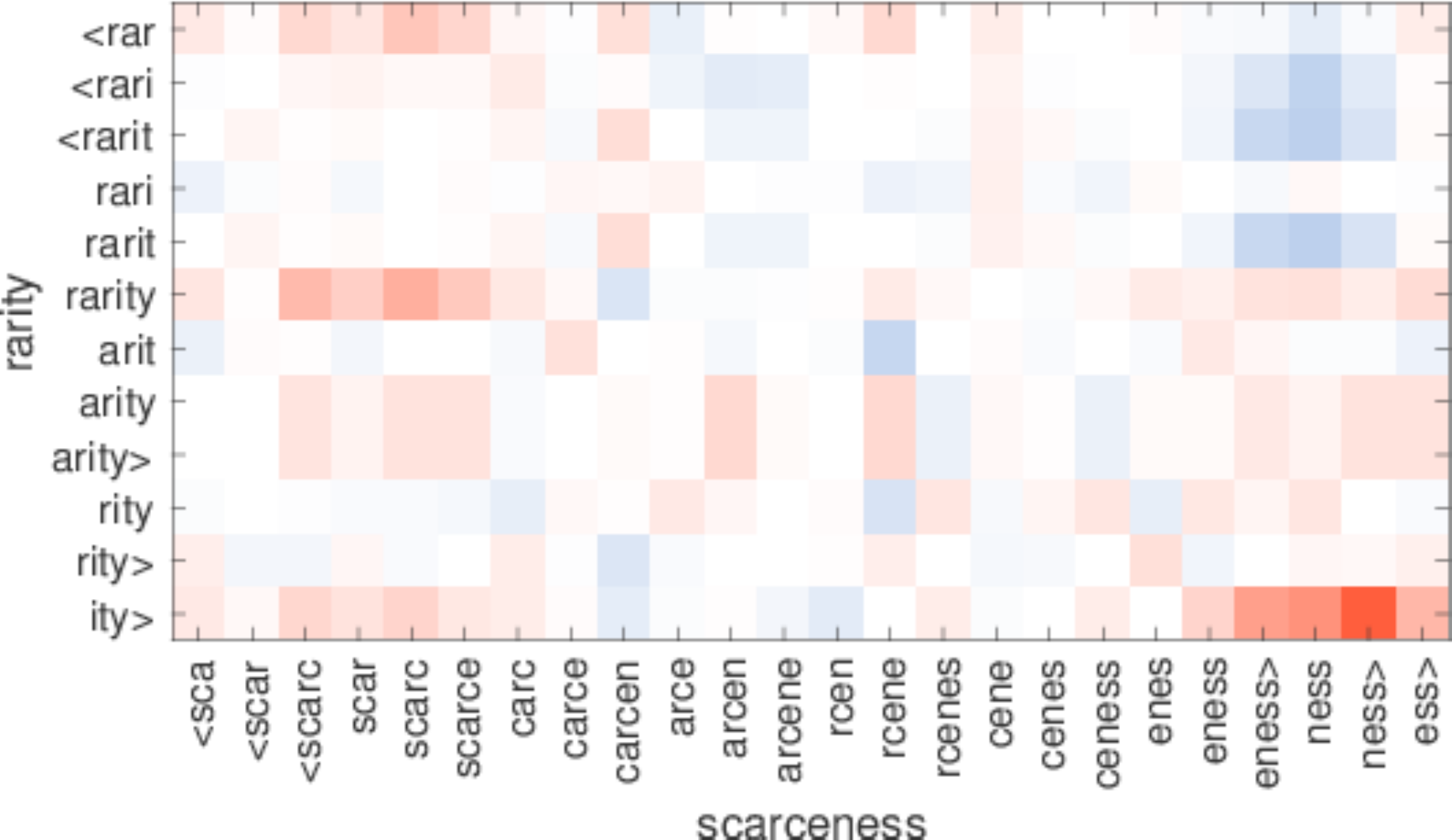}
    \vspace{0.5em}
    \\
    \includegraphics[width=\linewidth]{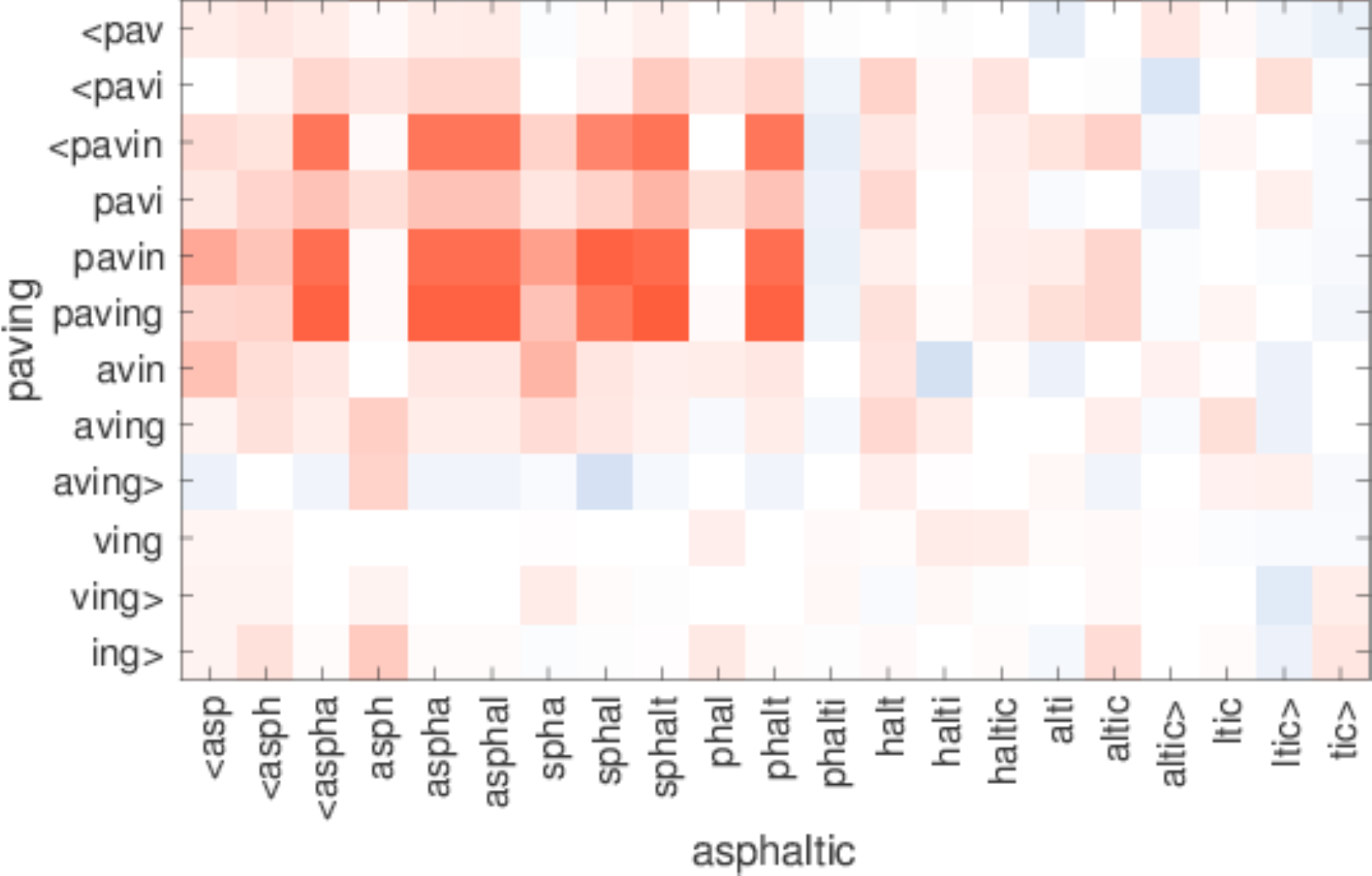}
    \vspace{0.5em}
    \\
    \includegraphics[width=\linewidth]{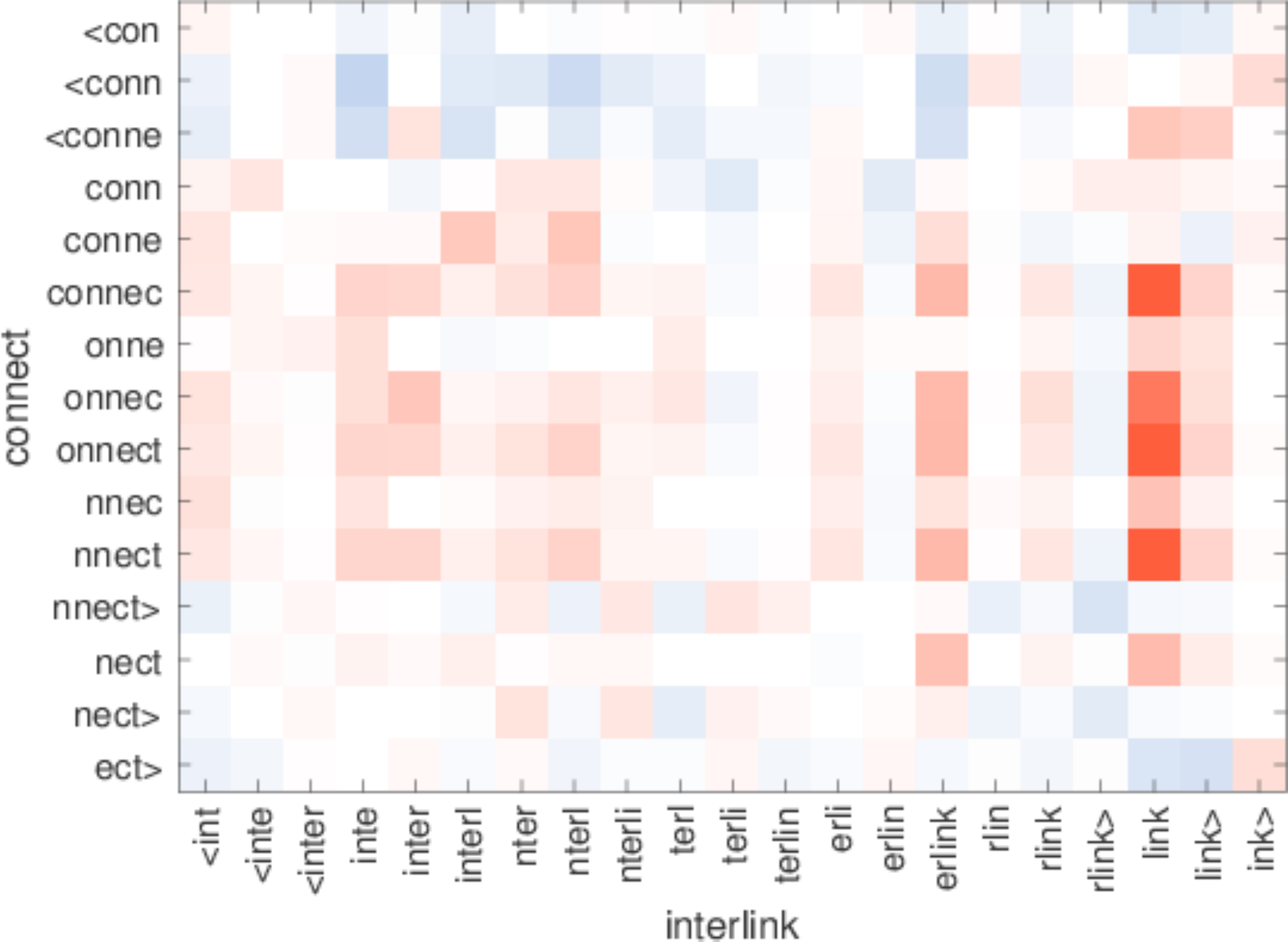}
  \end{minipage}
  \hspace{2.5em}
  \begin{minipage}[t]{.475\textwidth}
    \vspace{0pt}
    \includegraphics[width=\linewidth]{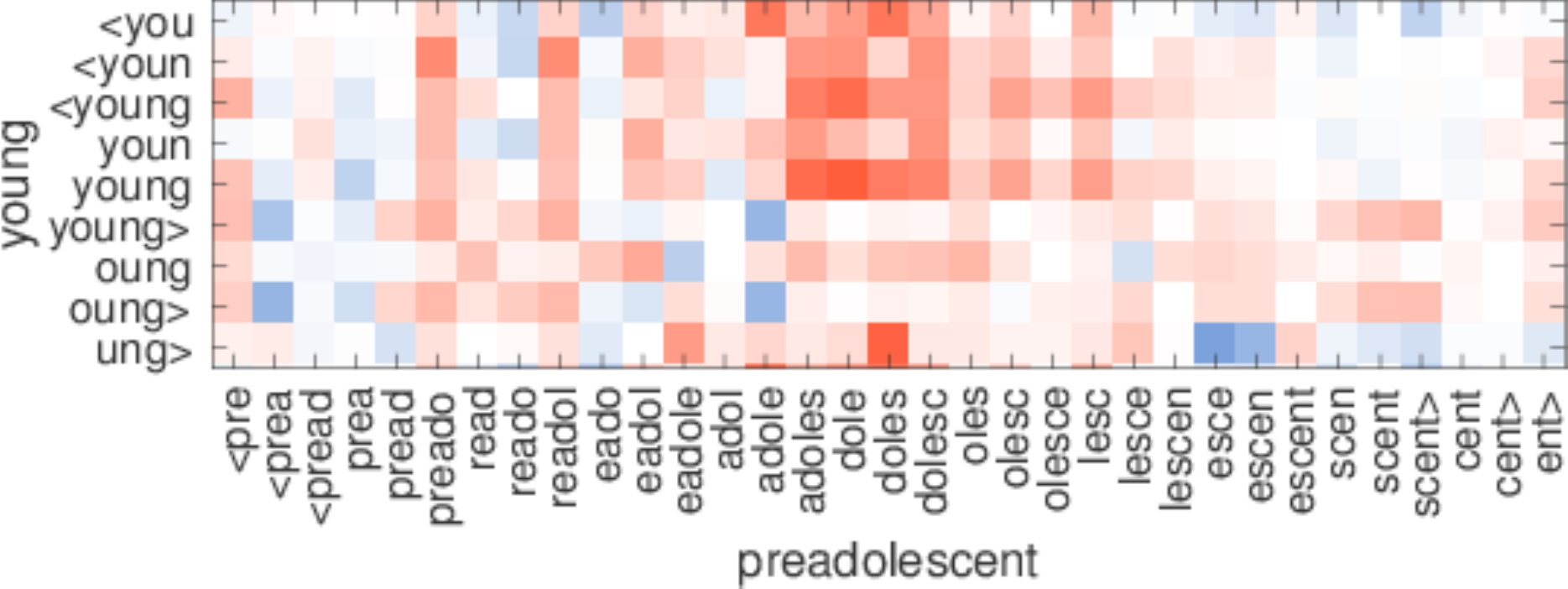}
    \vspace{0.5em}
    \\
    \includegraphics[width=\linewidth]{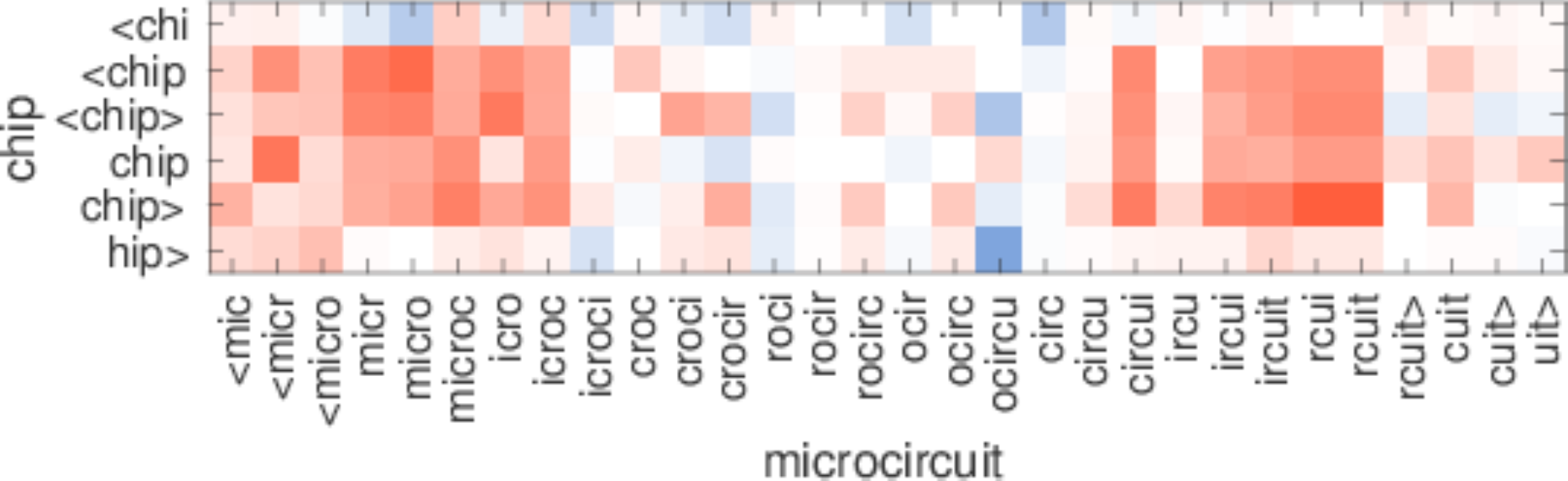}
    \vspace{0.5em}
    \\
    \includegraphics[width=\linewidth]{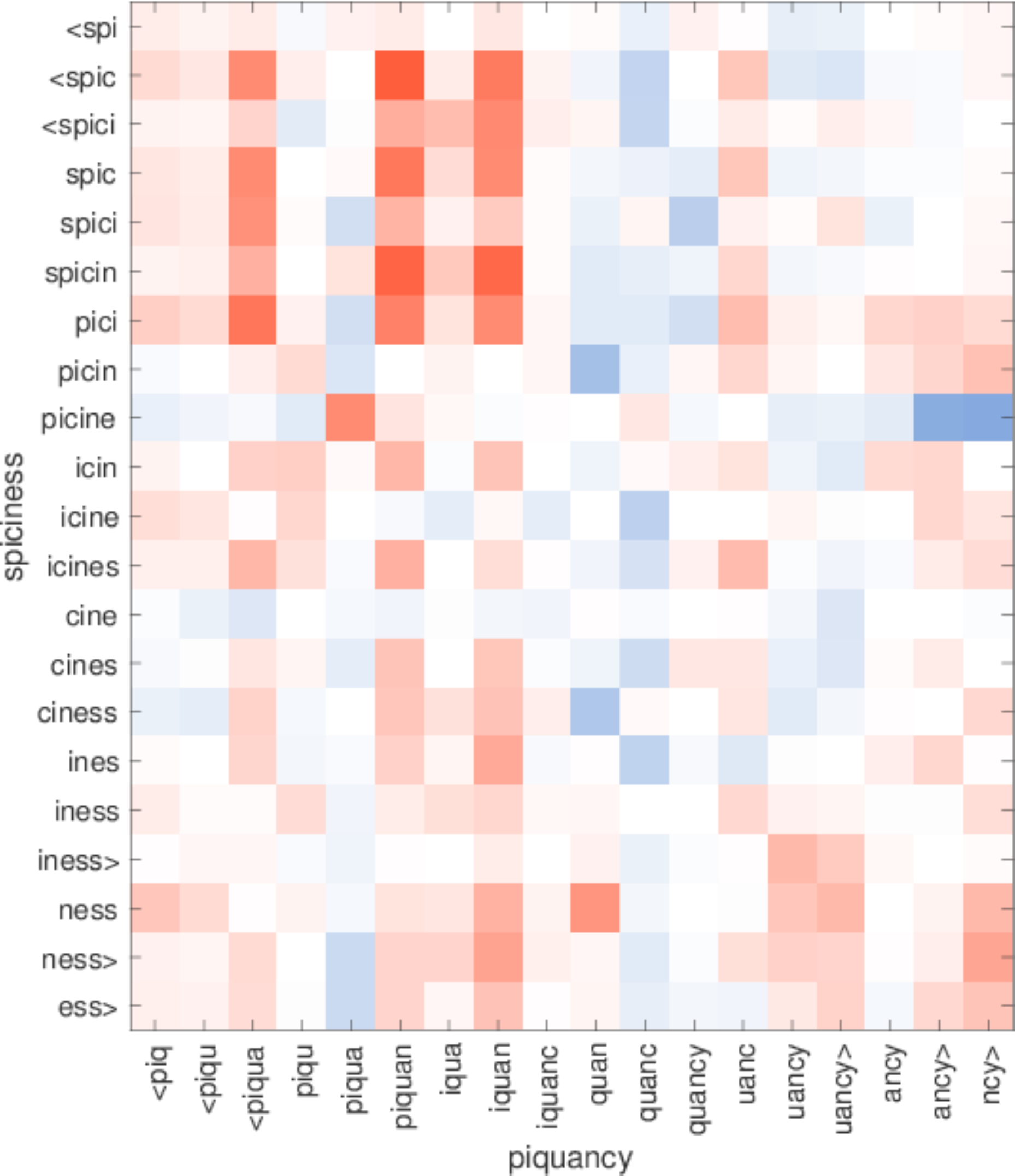}
  \end{minipage}
  \caption{
    Illustration of the similarity between character $n$-grams in out-of-vocabulary words.
    For each pair, only one word is OOV, and is shown on the $x$ axis.
    Red indicates positive cosine, while blue negative.
  }
  \label{fig:ngram-match}
\end{figure*}

As described in Sec.~\ref{sec:model-ngrams}, our model is capable of building word vectors for words that do not appear in the training set.
For such words, we simply average the vector representation of its $n$-grams.
In order to assess the quality of these representations, we analyze which of the $n$-grams match best for OOV words by selecting a few word pairs from the English RW similarity dataset.
We select pairs such that one of the two words is not in the training vocabulary and is hence only represented by its $n$-grams.
For each pair of words, we display the cosine similarity between each pair of $n$-grams that appear in the words.
In order to simulate a setup with a larger number of OOV words, we use models trained on~$1\%$ of the Wikipedia data as in Sec.~\ref{sec:exp-data-size}.
The results are presented in Fig.~\ref{fig:ngram-match}.

We observe interesting patterns, showing that subwords match correctly.
Indeed, for the word \emph{chip}, we clearly see that there are two groups of $n$-grams in \emph{microcircuit} that match well.
These roughly correspond to \emph{micro} and \emph{circuit}, and $n$-grams in between don't match well.
Another interesting example is the pair \emph{rarity} and \emph{scarceness}.
Indeed, \emph{scarce} roughly matches \emph{rarity} while the suffix \emph{-ness} matches \emph{-ity} very well.
Finally, the word \emph{preadolescent} matches \emph{young} well thanks to the \emph{-adolesc-} subword.
This shows that we build robust word representations where prefixes and suffixes can be ignored if the grammatical form is not found in the dictionary.
 
\begin{table*}[p]
  \centering
  \begin{tabular}{lcccccc}
    \toprule
    query             & tiling    & tech-rich        & english-born & micromanaging & eateries    & dendritic  \\
    \midrule
    \texttt{sisg}     & tile      & tech-dominated   & british-born & micromanage   & restaurants & dendrite   \\
                      & flooring  & tech-heavy       & polish-born  & micromanaged  & eaterie     & dendrites  \\
    \midrule
    \texttt{sg} & bookcases & technology-heavy & most-capped  & defang        & restaurants & epithelial \\
                      & built-ins & .ixic            & ex-scotland  & internalise   & delis       & p53        \\
    \bottomrule
  \end{tabular}
  \caption{
    Nearest neighbors of rare words using our representations and \texttt{skipgram}.
    These hand picked examples are for illustration.}
  \label{tab:nn}
\end{table*}

